\documentclass[conference]{IEEEtran}

\usepackage{graphicx}
\usepackage[center]{caption}
\usepackage{wrapfig}
\usepackage{textcomp}
\usepackage{mathrsfs}
\usepackage{esint}
\graphicspath{ {./images/} }
\providecommand{\keywords}[1]{\textbf{\textit{Keywords ---}} #1}

\begin{document}
\title{Distance Correlation Sure Independence Screening for Accelerated Feature Selection in Parkinson's Disease Vocal Data }
\author{
    \IEEEauthorblockN{Dan Schellhas\IEEEauthorrefmark{1},
        Bishal Neupane\IEEEauthorrefmark{2},
        Deepak Thammineni\IEEEauthorrefmark{3},
        Bhargav Kanumuri\IEEEauthorrefmark{4},
        and Robert C. Green II\IEEEauthorrefmark{5}
    }
    \IEEEauthorblockA{
        Department of Computer Science\\
        Bowling Green State University\\
        Bowling Green, Ohio, 43403\\
        dschell@bgsu.edu\IEEEauthorrefmark{1},
        bneupan@bgsu.edu\IEEEauthorrefmark{2},
        deepakt@bgsu.edu\IEEEauthorrefmark{3},
        bkanumu@bgsu.edu\IEEEauthorrefmark{4}, 
        greenr@bgsu.edu \IEEEauthorrefmark{5}
    }
}

\maketitle

\begin{abstract}
    With the abundance of machine learning methods available and the temptation of using them all in an ensemble method, having a model-agnostic method of feature selection is incredibly alluring. Principal component analysis was developed in 1901 and has been a strong contender in this role since, but in the end is an unsupervised method. It offers no guarantee that the features that are selected have good predictive power because it does not know what is being predicted. To this end, Peng et al. developed the minimum redundancy-maximum relevance (mRMR) method in 2005. It uses the mutual information not only between predictors but also includes the mutual information with the response in its calculation. Estimating mutual information and entropy tend to be expensive and problematic endeavors, which leads to excessive processing times even for dataset that is approximately 750 by 750 in a Leave-One-Subject-Out jackknife situation. To remedy this, we use a method from 2012 called Distance Correlation Sure Independence Screening (DC-SIS) which uses the distance correlation measure of Székely et al. to select features that have the greatest dependence with the response. We show that this method produces statistically indistinguishable results to the mRMR selection method on Parkinson's Disease vocal diagnosis data 90 times faster.
\end{abstract}
\section{Introduction}
Parkinson's Disease (PD) is a neurodegenerative motion disorder through which speech, voice, and body movements may malfunction. PD is one of the most common diseases after Alzheimer's spreading progressively over time in a human body \cite{compare}. Consequently, the disease can be diagnosed by estimating dysphonia produced by a subject who may have PD. Audio data is notoriously high dimensional. So, some forms of feature derivation and selection are required. Preferably, we would use a stable feature selection algorithm: one that selects the same kinds of features even if there is a dramatic change in which data points are available \cite{Gulgezen}. One such type of feature selection, Minimum Redundancy-Maximum Relevance (mRMR), is fairly commonly used for this task. Sakar et al.\cite{compare} used this method  alongside their tunable-Q wavelets since it provides reliable results regardless of model by estimating maximum joint dependency of features, eliminating the redundant features, and reducing the dimensionality. Features are ranked based upon their relevance with class labels and redundancy compared to other features.

In addition to the mutual information method employed by mRMR, distance correlation can also be considered for measuring the relevance of a feature. Distance correlation can use Euclidean distance, Manhattan distance, Minkowski distance, cosine similarity, or any other metric to measure the dependence between features, and to identify which are relevant. We can also greatly simplify the process by removing the minimum-redundancy constraint from mRMR and focus solely on maximum relevance. Li et al. call this method Distance Correlation Sure Independent Selection (DC-SIS) \cite{dcsis}. Once a set of maximally relevant features are selected, machine learning classifiers such as Random Forest, Support Vector Classifier (SVC), k-Nearest-Neighbor(k-NN), Multilayer perceptron, Logistic regression, Naive Bayes are applied on the new datasets. 

The main reason behind choosing the DC-SIS over mRMR is that mRMR requires too much processing time to perform feature selection on the PD dataset in \cite{compare} even when running on C code in parallel.

Performing the mRMR feature selection in a jackknife estimate of performance required 28 hours on 2.2 GHz Intel Core i7 processor using the \verb|pymrmr| Python package. Performing a comparable selection using DC-SIS required only 18 minutes, and produced predictions with performance not statistically distinguishable from mRMR.

The rest of the paper is presented as follows. In Section II, we shall describe the dataset that we used in our study and the related works. In Section III, we discuss our methodology and approach of feature selection. We present our results in Section IV. In Section V, we discuss the result, and implication of our study and potential future works. We conclude our study in Section VI.

\section{Background}
\subsection{Dataset Description}
The data for this experiment is the Parkinson's Disease Classification Data Set \cite{compare} from the UCI Machine Learning Repository that consists of 756 observations of 755 variables. The observations are from 188 patients  with PD, 107 male and 81 female, and their ages vary from 33 to 87. The control group contains 64 healthy individuals without PD, 23 male and 41 female, with ages between 41 and 82 \cite{compare}. Each subject contributes three observations to the dataset. Each observation is derived from the subject speaking, and holding, a vowel sound into a microphone. Gender is the only categorical predictor with the rest being quantitative coming from six large categories.
\subsubsection{Baseline features}
There are 21 baseline features that formed the initial research into diagnosing PD using vocal data. There are five features that measure the jitter: the instability of the fundamental frequency of the vocalization. Six features measure the shimmer: the instability of the amplitude of the vocalization. Five features characterize the fundamental frequency and its non-time-series statistical properties. The remaining features quantify the signal-to-noise ratio, signal entropy, self-similarity, and logarithmic entropy.

\subsubsection{Time-frequency features}
There are 11 features that deal with the timbre of the voice in the time-frequency domain. Three such features measure the intensity of the vocalization. Four measure the strength of the first four formats of the timbre. An additional four measure the spacing between the formats.

\subsubsection{Mel Frequency Cepstral Coefficients}
Eighty-four (84) features are used to measure non-vocal-fold effects. Vocalizations are complex physical processes and employ many muscles in their formation. These MFCC features attempt to characterize these effects that take place in other portions of the vocal tract than the folds.

\subsubsection{Wavelet Transform Features}
These 182 features are derived from an orthogonal wavelet transform that has been normalized to the fundamental frequency. Unlike the tunable Q wavelets of Sakar et al., these wavelets are general-purpose and form an orthogonal basis for representing the vocalization instead of an overcomplete one. This makes them easy to use, but potentially insufficient for inference.

\subsubsection{Vocal Fold Features}
These 22 vocal fold features provide a more detailed quantification of the baseline vocal fold features described above. The Glottis quotient, with 3 features, focuses on the periodicity of glottal closure. The Glottal-to-Noise Excitation, with 6 features, quantifies the signal-to-noise ratio more deeply. The Vocal Fold Excitation Ratio, with 7 features, also quantifies the signal-to-noise ratio but uses non-linear methods to do so. Lastly, the Empirical Mode Decomposition, with 6 features, uses elementary basis functions and measurements of entropy to decompose the signal similarly to a wavelet transform.

\subsubsection{Tunable-Q Wavelet Transform}
The remaining 434 features are derived using the tunable-Q wavelet transform suggested and implemented by Sakar et al. This transform is overcomplete and will consequently have redundancy in its results. However, this overcompleteness allows for more intuitive features after a selection process is used. Since the basis functions are uniformly spaced with respect to human experience, they also align well with respect to human vocalization. These features tend to be selected not only due to their number, but also from their predictive ability.

\subsection{Related works}
DC-SIS is compared directly to that of Sakar et al.\cite{compare}. In their study, they apply tunable Q-factor wavelet transform (TQWT) to the voice signals of PD patients and generate discriminating features from summary statistics of a bag-of-waves model. They combine several earlier feature sets and compares the performance of TQWT with them. The result of the study suggests the better performance of TQWT in terms of accuracy and complementary information in PD classification to combine feature selection and develop an improved system with an ensemble model.

With Sakar et al. are focused on their wavelets, their ensemble method still rests upon the features that it receives. These features are chosen via the work of Peng et al. \cite{feature} who investigate how to select excellent features according to maximal statistical dependency criterion based on mutual information. They derive an equivalent form of maximal dependency condition named minimum redundancy-maximum relevance (mRMR) for first-order incremental feature selection. Since mRMR is model agnostic, it can be combined with other feature selectors, which can significantly improve classification accuracy.

Some other studies that work with similar or identical data include, Polat \cite{smote} who proposes a new method with two phases to detect Parkinson's disease using the features obtained from speech signals using hybrid machine learning methods. Synthetic Minority Over-Sampling Technique (SMOTE) is used to transform imbalance data. Then they use Random Forests classification for classification of Parkinson's disease datasets to generate promising results.

John Wu \cite{john} uses a multilayer perceptron artificial network to train and classify PD patients and healthy individuals. They evaluate this algorithm on Parkinson's disease datasets by emphasizing the imbalanced nature.

Mostafa et al. \cite{mostafa} evaluated the performances of Decision Tree, Naive Bayes, and Neural Network classification methods for diagnosis of Parkinson's disease. They conclude that Decision Tree produces the highest accuracy (91.63\%), followed by Neural Network (91.01\%) and Naive Bayes (89.46\%).

Gunduz \cite{deep} proposes two frameworks based on Convolutional Neural Networks (CNN) to classify Parkinson's Disease using sets of vocal (speech) features. They use 9-layered CNN as inputs and pass the feature sets to parallel input layers. They extract in-depth features from these parallel branches that are effective in boosting up discriminative power of the classifiers.

Wang et al. \cite{xgboost} present a python package named Imbalance-XGBoost by combining XGBoost software with weighted and focal losses to deal with binary label-imbalanced classification tasks. They use unique XGBoost methods and offer state-of-the-art performances and focus on the new perspective to study the imbalanced datasets.

Cai et al. \cite{cai} proposed Chaotic Bacterial Foraging Optimization based on the fuzzy k-nearest neighbor (FKNN) method for early PD diagnosis.

Truncer and Dogan \cite{octopus} have access to the vocal signals themselves, and introduce a novel octopus based feature extraction method to introduce a general signal recognition method. They combined preprocessing, feature extraction, feature selection, classification, and post-processing phases to achieve a high success rate and low computational complexity.

Badem et al. \cite{badem} have a competing feature selection method based on the artificial Bee Colony metaheuristic to also improve the result with classification methods that we have used in this study.

\section{Methodology}
\subsection{Data Transformation}
To ensure the statistical guarantees of both the feature selection and the machine learning algorithms, the predictors are standardized before the features are selected. Despite the statistical benefits of standardizing the data, we also applied the analysis using predictor normalization and sample normalization. Neither method of normalization produced results were significantly different from standardization.

\subsection{Minimum Redundancy-Maximum Relevance}
In Sakar et al.\cite{compare}, the authors use mRMR as a model-agnostic method for feature selection to demonstrate the effectiveness of the TQWT for diagnosing PD patients from vowel vocalization data. This method, as presented by Peng et al. \cite{feature}, attempts to maximize the dependency of the response (having the disease) on the predictors. The mutual information definition of dependency has many attractive properties, but estimability is not one of them. Consequently, the authors solve a simpler, but related, problem: minimum redundancy-maximum relevance. Peng et al. \cite{feature} minimize the mutual information between predictors while simultaneously maximizing the mutual information between the predictors as a whole and the response.

The mutual information of random variables $X$ and $Y$, $I(X,Y)$ is defined in terms of their joint density function $f(X,Y)$ and marginal densities $f(X)$ and $f(Y)$:
\begin{equation}
    I(X,Y) = \iint f(X,Y) \log\frac{f(X,Y)}{f(X)f(Y)} dX dY
    \label{eq:i}
\end{equation}

From this, definition of mutual information is derived from the measure of feature relevance $D$ for feature $X_{\cdot i}$ within feature set $S$, and class $c$:
\begin{equation}
    D(S,c) = \frac{1}{|S|} \sum_{X_{\cdot i} \in S} I(X_{\cdot i}, c)
    \label{eq:d}
\end{equation}

Similarly, the redundancy of features totals the mutual information between features in the model:
\begin{equation}
    R(S) = \frac{1}{|S|^2} \sum_{X_{\cdot i},X_{\cdot j} \in S} I(X_{\cdot i}, X_{\cdot j})
    \label{eq:r}
\end{equation}

Thus, the mRMR statistic of a given feature set $\Phi$ can be defined as the difference between the relevance (\ref{eq:r}) and the redundancy (\ref{eq:d}):
\begin{equation}
    \Phi(D,R) = D - R
    \label{eq:phi}
\end{equation}

or, alternatively, as their quotient. This represents the difference of their logs:
\begin{equation}
    \Phi(D,R) = \frac{D}{R}
\end{equation}

\subsection{Minimum Redundancy-Maximum Relevance Computation}
To perform the mRMR feature selection, we used the \verb|pymrmr| package in Python, which uses C code via Cython to do the computations efficiently and in parallel. Estimating the mutual information of continuous data in (\ref{eq:i}) is a difficult and ongoing problem. Since it is a double integral (or reduced to a double sum), the computational complexity is typically some multiple of $O(N^2)$ depending on the way the included predictors are handled since $X$ and $Y$ could each be multivariate. This method of computing mutual information is then applied to computing the relevance (\ref{eq:d}) and redundancy (\ref{eq:r}).

Best subset selection using this criterion is as difficult as with other criteria, so a greedy forward stepwise selection is done instead. Since this method adds a single predictor at a time, we only need to compute the differential of (\ref{eq:phi}) to select the best change. Thus, computing the change in relevance due to adding a given predictor effectively amounts to computing the mutual information of that predictor with the response. 

Then, computing the change in redundancy involves finding the mutual information between the given predictor and the predictors already selected. The predictor with largest differential $\Phi$ is then added to the model. The complexity of this redundancy computation is likely $O(N^2 \cdot k)$ where $k$ is the current size of the model.  This method produces an ordered list of features which can be used to build a model of any size up to the user-chosen stopping size.

Although the asymptotic complexity is $O(N^2 \cdot p)$ for sample size $N$ and total number of features $p$, as the size of the current model increases, it becomes more expensive to compute each iteration causing the selection of 50 features in a Leave-One-Subject-Out setting to require 28 hours on 2.2 GHz Intel Core i7 processor using the \verb|pymrmr| package.

\subsection{Distance Correlation}
 
 In Székely et al. \cite{dcorr}, the authors present a measurement of dependence analogous to the Pearson correlation coefficient but instead of mere lack of correlation at a value of zero, it ensures independence instead. By measuring the distances (typically Euclidean) between data points and finding the correlation of these distances between two variates, distance correlation is able to make stronger statements about dependence than the Pearson $R$, the Kendall $\tau$, or the Spearman $\rho$, even when the two variates have different dimension.

For doubly centered distance matrices $A$ and $B$ of random variables or vectors $X$ and $Y$, the distance covariance is defined:
\begin{equation}
    \mathscr{V}^2_n(X,Y) = \frac{1}{n}\sum^n_{k,\ell=1}A_{k\ell}B_{k\ell}
    \label{eq:dcov}
\end{equation}
From this distance covariance, we can compute the distance correlation similarly to the computation of the Pearson correlation:
\begin{equation}
    \mathscr{R}^2(X,Y) = \frac{\mathscr{V}^2_n(X,Y)}{\sqrt{\mathscr{V}^2_n(X)}\sqrt{\mathscr{V}^2_n(Y)}}
    \label{eq:dcorr}
\end{equation}

\subsection{Distance Correlation Computation}
To compute the distance correlation, one must first be able to compute the distance covariance defined by (\ref{eq:dcov}). First, the pairwise distances between samples of the response need to be computed and placed into a matrix. Similarly, the pairwise distances between samples of the current predictor are placed into another matrix. The distance measure can be any that satisfies the triangle inequality, but Euclidean distance is typical. Since the distances are pairwise in the number of samples and involve only a single dimension at a time, the computational complexity of this step is $O(N^2)$.

The matrices are then doubly centered by removing the row mean from each row and the column mean from each column, then adding the grand mean back in. These matrices are then multiplied element-wise and the resulting matrix elements are summed. Both of these operations are also $O(N^2)$, so the overall complexity of computing the distance covariance is $O(N^2)$. The division by $n$ can be ignored as it is always cancelled out in the computation of the correlation.

To compute the distance correlation, we simply divide the distance covariance of the response with a given predictor by the square root of the distance covariance of the response with itself and by the square root of the distance covariance of the predictor with itself. On the surface, this would merely be doing three $O(N^2)$ operations with two square roots and two multiplications, but a significant amount of computation can be saved by sharing the doubly centered distance matrices. The overall complexity remains at $O(N^2)$, but there are computational savings to be done with a careful implementation.

\subsection{Distance Correlation Sure Independent Screening}

The method that we are proposing to use, Distance Correlation Sure Independent Screening (DC-SIS) \cite{dcsis}, uses the distance correlation between each predictor in isolation and the response as the importance of that predictor. Since the importance's are independent of one another, the feature selection process involves merely sorting the list and choosing the $k$ features with highest importance. The user is then able to either select an arbitrary number of features or combine this feature importance ordering with a model-specific stepwise selection algorithm.

This method is "naive" because the redundancies between predictors are ignored similarly to the Naive Bayes Classifier. This implies an assumption that the predictors are independent, and thus have a distance correlation among any set predictors of 0. In the language of \cite{feature}, by assuming the redundancy ($R$) is zero, we need to only optimize the relevance ($D$). Thus, our goal is to optimize a new criterion $\Phi^*$ as in (\ref{eq:phi}) with respect to a Naive Distance Correlation sense of redundancy $R^*$ and relevance $D^*$:
\begin{equation}
    \max \Phi^* = D^* - R^*
    \label{eq:phistar}
\end{equation}

Our independence assumption leads to $R^* = 0$, and our definition of relevance using distance correlation from (\ref{eq:dcorr}) implies the following:
\begin{equation}
    D^*(S,c) = \frac{1}{|S|}\sum_{X_{\cdot i} \in S} \mathscr{R}^2(X_{\cdot i}, c)
\end{equation}

\subsection{DC-SIS Computation}

To do this screening, we use the distance correlation algorithm described above between the response and each predictor independently. This provides a feature quality rating for each predictor with a computational complexity of $O(N^2)$. We can then sort the feature qualities and select the $k$ best using any standard sorting algorithm, which is $O(p \log p)$ for $p$ total predictors. Since the data is not modified in this process and the predictors are not compared against one another, the $p$ distance correlation computations can be computed in parallel. Since we must compute $p$ separate distance correlations, the computational complexity is $O(N^2 \cdot p)$, but sometimes can be saved by sharing the distance matrix of the response across the computations of each correlation. Similar savings could be exploited across folds in a jackknife setting, but this was not implemented.

The runtime advantage of DC-SIS may not be generalizable. However, One peculiar feature of this PD dataset is that it is only as long as it is wide. Consequently, methods that are bottle-necked by the number of features perform the same as those bottle-necked by the number of samples. Distance correlation has complexity $O(N^2)$, and is thus intended for small datasets where the asymptotic results provided by the central limit theorem do not apply. As the number of samples increases, this method quickly becomes untenable.

\subsection{Additional Datasets}
To further demonstrate the computational properties of mRMR and DC-SIS, we perform a similar feature selection on each fold of a jackknife for three additional datasets from the UCI Machine Learning Repository with a similar shape. First, the Data for Software Engineering Teamwork Assessment in Education Setting Data Set (SETAP) \cite{setap} has 74 observations of 84 predictors. Then, the Urban Land Cover Data Set (ULC) \cite{ulc} has 168 observations of 147 predictors. Lastly, the Arrhythmia Data Set (ARR) \cite{arr} has 452 observations of 279 predictors. Since these datasets are being used for computational benchmarking but not predictive analysis, their predictors and features will not be discussed.

\subsection{Predictive Models}
To demonstrate the efficacy of the DC-SIS method, we will be competing against the "All feature subsets" section of Table 5 from \cite{compare}. In this portion of the table, Sakar et al. demonstrated the accuracy, F1 score, and Matthews correlation coefficient (MCC) of Naive Bayes, Logistic Regression, k-Nearest Neighbor, Multilayer Perceptron, Random Forest, and Support Vector Classifiers with linear and radial kernels with features selected from all of those available.  

We will now describe the settings used for each machine learning model. Whenever possible, a grid search was used to optimize the hyperparamters. Since Naïve Bayes has no hyperparameters other than the prior probability of each class, we used the default behavior or inferring the prior from the training data. For logistic regression, we used the \verb|saga| solver with the elastic net penalty at an L1 ratio of 0.25. For k-NN, we used 18 neighbors and the L3 distance. For Multilayer Perceptron, we employed hidden layers with 50, 25, 10, and 5 nodes respectively, the ReLU activation function and an initial learning rate of 0.01. For the Random Forest classifier, we allowed 200 trees with a maximum of 20 leaf nodes and weighted each observation to balance the classes. For the linear Support Vector Classifier we set a cost of 0.5, and 0.8 for the radial basis function version.

\subsection{Validation}
As in \cite{compare}, we measure the performance of each model via a Leave-One-Subject-Out jackknife estimate of accuracy, F1 score, and the Matthews correlation coefficient. One subject (three observations) is removed from the dataset and held in reserve. Each model is trained on the remaining data and then tested on the subject held in reserve. This is repeated for each subject in turn until we have as many predictions as we have samples.

Ordinary bootstrap and jackknife estimates of performance cannot be used since there are three measurements per subject and treating them as independent observations would cause information to be leaked into each training/bag set from each validation set or out-of-bag set. So, long as all three measurements per subject remain together, bootstrapping and k-fold cross-validation can occur. 

However, the typical 5-fold cross-validation reduces the training set size by 20\% and bootstrapping reduces its information content by 36\% on average. Both of these reductions are too much for data this small, and cause the performance of all models to drop to insignificant levels. As such, we also use a Leave-One-Subject-Out jackknife estimate for the standard error of accuracy, but this precludes estimating the standard error for the F1 score, and the MCC. 
 
Any transformation of the data must be done within each fold of the jackknife to prevent any information leakage. This includes not just the training of the models but under- or over-sampling, feature selection, or hyperparameter tuning. As we are focused on the feature selection, we did not implement any data balancing like in \cite{smote}. 

\subsection{Prediction}
Instead of treating each observation independently, the three measurements per subject are aggregated for each prediction. The three measurements are each classified, and the more common class is selected as the prediction for all three observations. This causes a small increase in prediction variance in exchange for a small decrease in the prediction bias.

\subsection{Model Shrinkage}
Since the choice of 50 features appears to be relatively arbitrary, we test DC-SIS on the full size 50 model and search for an optimal shrunken model. To do so, we can simply scan the performance of each size model between 1 and 50 and keep the smallest model that reaches a certain threshold. The threshold we have chosen is having estimated accuracy one standard error or less below the predicted accuracy of the size 50 model.

\section{Results}
\subsection{Prediction}
Using the same number of features as \cite{compare}, DC-SIS was able to provide comparable predictive performance to the much more expensive mRMR selection as shown in Table \ref{tab:acc}. The highest accuracy (0.86), F1 score (0.91), and Matthews correlation (0.60) were all achieved by the random forest.

\begin{figure}
    \centering
    \includegraphics[width=0.54\textwidth, height=6cm]{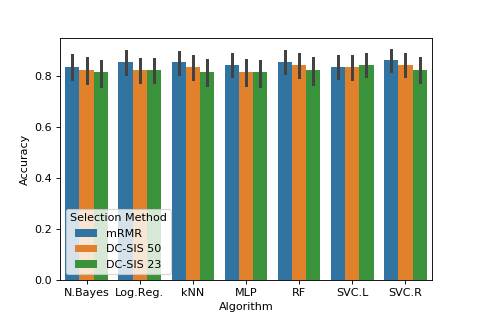}
    \caption{Comparison of accuracy on different feature selection methods }
    \label{fig:acc.png}
\end{figure}

Since generating the complete selection path with DC-SIS is as easy as computing a single selection, we also searched for the smallest model that had accuracy within one standard error ($\approx 0.01$) of the size 50 model. On average, this occurred at 23 features. The results of these reduced models are also included in Table \ref{tab:acc}. As such, we can cut the model in half and still achieve results that are not statistically different from those achieved by mRMR as shown in Figure \ref{fig:acc.png}. This is exemplified by the linear SVC that still achieved an accuracy of 0.84, F1 score of 0.90, and a Matthews Correlation of 0.54.

\begin{table*}[]
    \caption{Results obtained from top 50 and top 23 features selected by mRMR and DC-SIS}
    \label{tab:acc}
    \begin{tabular}{|l|l|l|l||l|l|l||l|l|l|}
        \hline  & \multicolumn{3}{c||}{\textbf{Accuracy}}
                & \multicolumn{3}{c||}{\textbf{F1-Score}}
                & \multicolumn{3}{c|}{\textbf{MCC}} \\ 
        \hline  & \textbf{mRMR}
                & \textbf{DC-SIS 50} 
                & \textbf{DC-SIS 23} 
                & \textbf{mRMR}
                & \textbf{DC-SIS 50} 
                & \textbf{DC-SIS 23} 
                & \textbf{mRMR}
                & \textbf{DC-SIS 50} 
                & \textbf{DC-SIS 23} \\ 
        \hline \textbf{Naive Bayes} 
                & 0.83 & 0.82 & 0.81 & 0.83 & 0.88 & 0.88 & 0.54 & 0.50 & 0.47 \\ 
        \hline \textbf{Logistic regression}                         
                & 0.85 & 0.82 & 0.82 & 0.84 & 0.87 & 0.88 & 0.57 & 0.49 & 0.47 \\ 
        \hline \textbf{k-NN}                                        
                & 0.85 & 0.84 & 0.82 & 0.82 & 0.90 & 0.88 & 0.56 & 0.51 & 0.46 \\ 
        \hline \textbf{Multilayer perceptron} 
                & 0.84 & 0.81 & 0.81 & 0.83 & 0.87 & 0.88 & 0.54 & 0.46 & 0.46 \\ 
        \hline \textbf{Random Forest}
                & 0.85 & 0.84 & 0.82 & 0.84 & 0.90 & 0.89 & 0.57 & 0.54 & 0.50 \\ 
        \hline \textbf{SVC (Linear)}
                & 0.83 & 0.83 & 0.84 & 0.82 & 0.90 & 0.90 & 0.52 & 0.52 & 0.54 \\ 
        \hline \textbf{SVC (RBF)}
                & 0.86 & 0.84 & 0.82 & 0.84 & 0.90 & 0.89 & 0.59 & 0.53 & 0.48 \\ 
        \hline
    \end{tabular}
\end{table*}

\subsection{Performance}
By ignoring redundancy, DC-SIS does not get more expensive as the number of selected features grows. Since each predictor is treated as independent, selecting fifty features requires the same amount of processing as selecting two. By computing the correlations in parallel and sharing the response distance matrix across correlations, DC-SIS is able to rank all 755 predictors across all 252 folds in 18 minutes. It requires 28 hours for mRMR to rank only the top 50 predictors across the 252 folds. This is a 90-fold increase in speed with no significant difference in predictive power.

Performing the mRMR and DC-SIS feature selection on the SETAP, ULC, and ARR datasets demonstrated a comparable difference in runtime as observed with the Parkinson's data. Running mRMR on SETAP ($74 \times 84$) required 101 seconds while DC-SIS only required 0.72 for a 141-fold speed-up. For ULC ($168 \times 147$), mRMR took 19 minutes and DC-SIS only 4.6 seconds for a 251-fold speed-up. The Arrythmia data ($452 \times 279$) needed 5 hours for mRMR to run but only 4 minutes for DC-SIS, which is a 77-fold speed-up. This shows that the 60-fold improvement on the Parkinson's data is not an anomaly and that the improvement amount varies depending on the size and shape of the data.

\subsection{Features}
Despite having indistinguishable performance in the prediction task, mRMR and DC-SIS do not select the same features. The difference in feature selection for each method caused by the jackknife is minimal, typically only one to four features change from fold to fold. However, between methods less than half of the features are shared for any given fold. The details for each dataset can be observed in Table \ref{tab:feat}.

For the PD dataset, mRMR has 35 features that were selected in every fold (for details, see Appendix). Similarly, DC-SIS has 42 features that were selected in every fold of the Leave-One-Subject-Out jackknife. Of these, the two procedures only agree on 12 features which are as follows:\\
\\
\verb|    tqwt_stdValue_dec_12| \\
\verb|    tqwt_TKEO_std_dec_12| \\
\verb|    tqwt_entropy_log_dec_12| \\
\verb|    tqwt_entropy_shannon_dec_11| \\
\verb|    mean_MFCC_2nd_coef| \\
\verb|    tqwt_TKEO_std_dec_11| \\
\verb|    std_delta_delta_log_energy| \\
\verb|    std_9th_delta_delta| \\
\verb|    std_8th_delta_delta| \\
\verb|    tqwt_energy_dec_12| \\
\verb|    std_7th_delta_delta| \\
\verb|    std_6th_delta_delta| \\

Of these 12, six are tunable-Q wavelet properties, five are standard wavelet properties, and one is a Mel  Frequency Cepstral Coefficient. These twelve are the most crucial for making predictions, but are insufficient on their own. DC-SIS and mRMR disagree on how to augment these twelve for the best predictions but end up with the same predictive power.

\begin{table}[]
    \caption{Percentage of features shared within and between selection methods}
    \label{tab:feat}
    \begin{tabular}{|l|l|l|l|}
        \hline  & \textbf{mRMR}
                & \textbf{DC-SIS} 
                & \textbf{Between} \\ 
        \hline \textbf{PD} 
                & 0.96 & 0.98 & 0.42 \\ 
        \hline \textbf{ARR}                         
                & 0.97 & 0.99 & 0.43 \\ 
        \hline \textbf{ULC}                                        
                & 0.93 & 0.98 & 0.45 \\ 
        \hline \textbf{SETAP} 
                & 0.95 & 0.94 & 0.55 \\ 
        \hline
    \end{tabular}
\end{table}

\section{Discussion}

Using DC-SIS is not meant as a global replacement for mRMR and other entropy-based feature selectors. We present it as a significantly more economical option when initially exploring the data. Since it can rank all of the features around 100 times faster than mRMR and can rank the top 50, DC-SIS can be used to find a reasonable upper bound for other, more expensive, feature selection methods. There is no sense in using mRMR to find the top 50 features in 28 hours, when you can show that only 23 are required using DC-SIS.

As future work, we can compare this performance with several other preprocessing techniques such as Principal Component Analysis (PCA) and Independent Component Analysis (ICA). We can also investigate  whether the predictive performance of DC-SIS generalizes to other Machine Learning Methods with datasets of different shapes. It is also possible to develop a distance correlation selection algorithm which takes into account feature redundancy. It could use stepwise selection like mRMR or some metaheuristic optimization method. Ordinary resampling like SMOTE \cite{smote} can offset the effects of imbalanced data, but clever resampling and combinations of methods can be applied to eliminate the effects of imbalanced datasets altogether \cite{handle}.

\section{Conclusion}
In this study, we used Distance Correlation Sure Independence Screening for feature selection, which exploits the relatively few samples relative to predictors. Using a Parkinson's disease remote diagnosis dataset, we achieved a comparable predictive performance to Minimum Redundancy-Maximum Relevance selection, but 90 times faster. The predictions were made using seven machine learning methods on a size 50 feature set within a Leave-One-Subject-Out jackknife. Being able to make high quality predictions with so much less computational load and time can save a great deal when the dataset has relatively few samples, but a great deal of predictors.

\bibliographystyle{ieeetr}
\bibliography{main}

\begin{thebibliography}{10}

\bibitem{compare}
C.~O. Sakar, G.~Serbes, A.~Gunduz, H.~C. Tunc, H.~Nizam, B.~E. Sakar,
  M.~Tutuncu, T.~Aydin, M.~E. Isenkul, and H.~Apaydin, ``A comparative analysis
  of speech signal processing algorithms for parkinson’s disease
  classification and the use of the tunable q-factor wavelet transform,'' {\em
  Applied Soft Computing}, vol.~74, pp.~255--263, 2019.

\bibitem{Gulgezen}
G.~{Gulgezen}, Z.~{Cataltepe}, and L.~{Yu}, ``Stable feature selection using
  mrmr algorithm,'' in {\em 2009 IEEE 17th Signal Processing and Communications
  Applications Conference}, pp.~596--599, April 2009.

\bibitem{dcsis}
R.~Li, W.~Zhong, and L.~Zhu, ``Feature screening via distance correlation
  learning,'' {\em Journal of the American Statistical Associationg}, vol.~107,
  no.~249, pp.~1129--1139, 2012.

\bibitem{feature}
{Hanchuan Peng}, {Fuhui Long}, and C.~{Ding}, ``Feature selection based on
  mutual information criteria of max-dependency, max-relevance, and
  min-redundancy,'' {\em IEEE Transactions on Pattern Analysis and Machine
  Intelligence}, vol.~27, pp.~1226--1238, August 2005.

\bibitem{smote}
K.~Polat, ``A hybrid approach to parkinson disease classification using speech
  signal: The combination of smote and random forests,'' in {\em 2019
  Scientific Meeting on Electrical-Electronics \& Biomedical Engineering and
  Computer Science (EBBT)}, (Istanbul, Turkey), pp.~1--3, April 2009.

\bibitem{john}
W.., ``Application of artificial neural network on speech signal features for
  parkinson’s disease classification,'' Master's thesis, California State
  University San Marcos, The address of the publisher, May 2019.

\bibitem{mostafa}
S.~A. Mostafa, A.~Mustapha, S.~H. Khaleefah, M.~S. Ahmad, and M.~A. Mohammed,
  ``Evaluating the performance of three classification methods in diagnosis of
  parkinson's disease,'' in {\em Recent Advances on Soft Computing and Data
  Mining} (R.~Ghazali, M.~M. Deris, N.~M. Nawi, and J.~H. Abawajy, eds.),
  (Cham), pp.~43--52, Springer International Publishing, 2018.

\bibitem{deep}
H.~Gunduz, ``Deep learning-based parkinson's disease classification using vocal
  feature sets,'' {\em {IEEE} Access}, vol.~7, pp.~115540--115551, 2019.

\bibitem{xgboost}
C.~Wang, C.~Deng, and S.~Wang, ``Imbalance-xgboost: Leveraging weighted and
  focal losses for binary label-imbalanced classification with xgboost,'' {\em
  CoRR}, vol.~abs/1908.01672, 2019.

\bibitem{cai}
Z.~Cai, J.~Gu, C.~Wen, D.~Zhao, C.~Huang, H.~Huang, C.~Tong, J.~Li, and
  H.~Chen, ``An intelligent parkinson’s disease diagnostic system based on a
  chaotic bacterial foraging optimization enhanced fuzzy knn approach,'' {\em
  Computational and Mathematical Methods in Medicine}, vol.~2018, pp.~1--24,
  2018.

\bibitem{octopus}
T.~Tuncer and S.~Dogan, ``A novel octopus based parkinson’s disease and
  gender recognition method using vowels,'' {\em Applied Acoustics}, vol.~155,
  pp.~75 -- 83, 2019.

\bibitem{badem}
H.~Badem, D.~Turkusagi, A.~Caliskan, and Z.~A. Cil, ``Feature selection based
  on artificial bee colony for parkinson disease diagnosis,'' in {\em 2019
  Medical Technologies Congress (TIPTEKNO)}, (Izmir, Turkey), pp.~1--4, October
  2019.

\bibitem{dcorr}
G.~J. Székely, M.~L. Rizzo, and N.~K. Bakirov, ``Measuring and testing
  dependence by correlation of distances,'' {\em The Annals of Statistics},
  vol.~35, no.~6, pp.~2769--2794, 2007.

\bibitem{setap}
D.~Petkovic, M.~Sosnick{-}P{\'{e}}rez, K.~Okada, R.~Todtenhoefer, S.~Huang,
  N.~Miglani, and A.~Vigil, ``Using the random forest classifier to assess and
  predict student learning of software engineering teamwork,'' in {\em 2016
  {IEEE} Frontiers in Education Conference, {FIE}}, pp.~1--7, {IEEE} Computer
  Society, 2016.

\bibitem{ulc}
B.~Johnson and Z.~Xie, ``Classifying a high resolution image of an urban area
  using super-object information,'' {\em ISPRS Journal of Photogrammetry and
  Remote Sensing}, vol.~83, pp.~40 -- 49, 2013.

\bibitem{arr}
H.~A. Guvenir, B.~Acar, G.~Demiroz, and A.~Cekin, ``A supervised machine
  learning algorithm for arrhythmia analysis,'' in {\em Computers in
  Cardiology}, pp.~433--436, 1997.

\bibitem{handle}
S.~B. Kotsiantis, D.~Kanellopoulos, and P.~E. Pintelas, ``Handling imbalanced
  datasets: A review,'' 2006.

\end{thebibliography}

\newpage
\appendix[Probability of Parkinson's Disease Features Being Selected by mRMR]
    \begin{tabular}{|l|l|}
        \hline
        \textbf{Feature} & \textbf{p} \\
        \hline
        std\_6th\_delta\_delta & 1.0 \\
        mean\_MFCC\_2nd\_coef & 1.0 \\
        tqwt\_kurtosisValue\_dec\_26 & 1.0 \\
        maxIntensity & 1.0 \\
        tqwt\_entropy\_shannon\_dec\_11 & 1.0 \\
        tqwt\_kurtosisValue\_dec\_36 & 1.0 \\
        tqwt\_stdValue\_dec\_33 & 1.0 \\
        std\_8th\_delta & 1.0 \\
        tqwt\_kurtosisValue\_dec\_20 & 1.0 \\
        tqwt\_entropy\_log\_dec\_27 & 1.0 \\
        tqwt\_TKEO\_std\_dec\_12 & 1.0 \\
        std\_delta\_delta\_log\_energy & 1.0 \\
        std\_11th\_delta\_delta & 1.0 \\
        f1 & 1.0 \\
        tqwt\_kurtosisValue\_dec\_27 & 1.0 \\
        tqwt\_entropy\_shannon\_dec\_32 & 1.0 \\
        tqwt\_energy\_dec\_12 & 1.0 \\
        std\_4th\_delta & 1.0 \\
        tqwt\_TKEO\_std\_dec\_35 & 1.0 \\
        tqwt\_entropy\_log\_dec\_12 & 1.0 \\
        tqwt\_skewnessValue\_dec\_25 & 1.0 \\
        std\_7th\_delta\_delta & 1.0 \\
        tqwt\_energy\_dec\_15 & 1.0 \\
        tqwt\_TKEO\_std\_dec\_11 & 1.0 \\
        tqwt\_skewnessValue\_dec\_36 & 1.0 \\
        tqwt\_skewnessValue\_dec\_26 & 1.0 \\
        tqwt\_kurtosisValue\_dec\_18 & 1.0 \\
        std\_8th\_delta\_delta & 1.0 \\
        tqwt\_energy\_dec\_11 & 1.0 \\
        tqwt\_entropy\_shannon\_dec\_33 & 1.0 \\
        tqwt\_stdValue\_dec\_12 & 1.0 \\
        apq11Shimmer & 1.0 \\
        std\_6th\_delta & 1.0 \\
        tqwt\_TKEO\_mean\_dec\_16 & 1.0 \\
        std\_9th\_delta\_delta & 1.0 \\
        tqwt\_kurtosisValue\_dec\_16 & 0.996 \\
        app\_LT\_entropy\_log\_5\_coef & 0.996 \\
        tqwt\_TKEO\_std\_dec\_5 & 0.992 \\
        ppq5Jitter & 0.992 \\
        tqwt\_entropy\_shannon\_dec\_8 & 0.988 \\
        tqwt\_TKEO\_mean\_dec\_8 & 0.984 \\
        tqwt\_entropy\_shannon\_dec\_34 &  0.98 \\
        GNE\_NSR\_SEO & 0.972 \\
        tqwt\_stdValue\_dec\_11 & 0.964 \\
        tqwt\_TKEO\_std\_dec\_6 &  0.96 \\
        tqwt\_energy\_dec\_27 & 0.932 \\
        tqwt\_kurtosisValue\_dec\_28 & 0.9 \\
        \hline
    \end{tabular}

    \begin{tabular}{|l|l|}
        \hline
        \textbf{Feature} & \textbf{p} \\
        
        \hline
        tqwt\_minValue\_dec\_11 & 0.892 \\
        b1 & 0.657 \\
        meanIntensity & 0.554 \\
        tqwt\_TKEO\_mean\_dec\_33 & 0.386 \\
        tqwt\_entropy\_log\_dec\_26 & 0.335 \\
        tqwt\_maxValue\_dec\_11 & 0.076 \\
        tqwt\_skewnessValue\_dec\_30 & 0.076 \\
        tqwt\_TKEO\_std\_dec\_19 & 0.048 \\
        tqwt\_energy\_dec\_33 & 0.036 \\
        tqwt\_energy\_dec\_25 & 0.032 \\
        tqwt\_maxValue\_dec\_12 & 0.028 \\
        DFA & 0.028 \\
        tqwt\_TKEO\_mean\_dec\_12 & 0.016 \\
        tqwt\_meanValue\_dec\_8 & 0.016 \\
        mean\_MFCC\_6th\_coef & 0.012 \\
        tqwt\_skewnessValue\_dec\_24 & 0.012 \\
        f2 & 0.012 \\
        app\_entropy\_log\_1\_coef & 0.004 \\
        tqwt\_TKEO\_std\_dec\_13 & 0.004 \\
        tqwt\_TKEO\_mean\_dec\_11 & 0.004 \\
        tqwt\_meanValue\_dec\_36 & 0.008 \\
        tqwt\_skewnessValue\_dec\_3 & 0.004 \\
        tqwt\_skewnessValue\_dec\_10 & 0.004 \\
        locAbsJitter & 0.004 \\
        GNE\_SNR\_TKEO & 0.004 \\
        std\_10th\_delta\_delta & 0.004 \\
        tqwt\_entropy\_log\_dec\_33 & 0.008 \\
        tqwt\_medianValue\_dec\_22 & 0.004 \\
        tqwt\_energy\_dec\_26 & 0.004 \\
        tqwt\_TKEO\_std\_dec\_33 & 0.004 \\
        tqwt\_stdValue\_dec\_29 & 0.004 \\
        tqwt\_TKEO\_mean\_dec\_6 & 0.008 \\
        tqwt\_energy\_dec\_7 & 0.004 \\
        tqwt\_meanValue\_dec\_33 & 0.004 \\
        RPDE & 0.004 \\
        tqwt\_TKEO\_mean\_dec\_34 & 0.004 \\
        tqwt\_entropy\_log\_dec\_28 & 0.004 \\
        tqwt\_TKEO\_mean\_dec\_30 & 0.004 \\
        std\_11th\_delta & 0.004 \\
        tqwt\_maxValue\_dec\_2 & 0.004 \\
        tqwt\_TKEO\_mean\_dec\_15 & 0.004 \\
        tqwt\_kurtosisValue\_dec\_21 & 0.008 \\
        tqwt\_energy\_dec\_10 & 0.004 \\
        tqwt\_medianValue\_dec\_34 & 0.004 \\
        tqwt\_energy\_dec\_2 & 0.004 \\
        \hline
    \end{tabular}

\appendix[Probability of Parkinson's Disease Features Being Selected by DC-SIS]
\begin{tabular}{|l|l|}
        \hline
        \textbf{Feature} & \textbf{p} \\
        
        \hline
        tqwt\_stdValue\_dec\_12 & 1.0 \\
        tqwt\_TKEO\_std\_dec\_12 & 1.0 \\
        tqwt\_entropy\_shannon\_dec\_12 & 1.0 \\
        tqwt\_TKEO\_mean\_dec\_12 & 1.0 \\
        tqwt\_entropy\_log\_dec\_12 & 1.0 \\
        tqwt\_maxValue\_dec\_12 & 1.0 \\
        tqwt\_minValue\_dec\_12 & 1.0 \\
        tqwt\_stdValue\_dec\_11 & 1.0 \\
        tqwt\_stdValue\_dec\_13 & 1.0 \\
        tqwt\_entropy\_shannon\_dec\_11 & 1.0 \\
        tqwt\_TKEO\_std\_dec\_13 & 1.0 \\
        mean\_MFCC\_2nd\_coef & 1.0 \\
        tqwt\_entropy\_shannon\_dec\_13 & 1.0 \\
        tqwt\_TKEO\_std\_dec\_11 & 1.0 \\
        tqwt\_maxValue\_dec\_13 & 1.0 \\
        tqwt\_minValue\_dec\_13 & 1.0 \\
        std\_delta\_delta\_log\_energy & 1.0 \\
        std\_9th\_delta\_delta & 1.0 \\
        std\_8th\_delta\_delta & 1.0 \\
        std\_delta\_log\_energy & 1.0 \\
        tqwt\_TKEO\_mean\_dec\_13 & 1.0 \\
        tqwt\_energy\_dec\_12 & 1.0 \\
        tqwt\_maxValue\_dec\_11 & 1.0 \\
        std\_7th\_delta\_delta & 1.0 \\
        tqwt\_entropy\_log\_dec\_11 & 1.0 \\
        std\_6th\_delta\_delta & 1.0 \\
        tqwt\_minValue\_dec\_11 & 1.0 \\
        tqwt\_entropy\_log\_dec\_13 & 1.0 \\
        tqwt\_TKEO\_mean\_dec\_11 & 1.0 \\
        std\_9th\_delta & 1.0 \\
        tqwt\_kurtosisValue\_dec\_26 & 1.0 \\
        tqwt\_kurtosisValue\_dec\_27 & 1.0 \\
        std\_10th\_delta\_delta & 1.0 \\
        std\_8th\_delta & 1.0 \\
        tqwt\_energy\_dec\_11 & 1.0 \\
        tqwt\_TKEO\_std\_dec\_14 & 1.0 \\
        std\_11th\_delta\_delta & 1.0 \\
        tqwt\_stdValue\_dec\_14 & 1.0 \\
        std\_7th\_delta & 1.0 \\
        tqwt\_entropy\_shannon\_dec\_16 & 1.0 \\
        tqwt\_entropy\_shannon\_dec\_14 & 1.0 \\
        tqwt\_stdValue\_dec\_16' & 1.0 \\
        \hline
    \end{tabular}    

\begin{tabular}{|l|l|}
        \hline
        \textbf{Feature} & \textbf{p} \\
        
        \hline
        std\_6th\_delta & 0.996 \\
        tqwt\_energy\_dec\_13 & 0.992 \\
        tqwt\_entropy\_log\_dec\_16 & 0.992 \\
        tqwt\_TKEO\_mean\_dec\_16 & 0.992 \\
        tqwt\_stdValue\_dec\_15 & 0.992 \\
        tqwt\_maxValue\_dec\_14 & 0.888 \\
        tqwt\_minValue\_dec\_14 & 0.622 \\
        tqwt\_entropy\_shannon\_dec\_15 & 0.928 \\
        tqwt\_stdValue\_dec\_7' &0.315, \\
        std\_10th\_delta & 0.195 \\
        tqwt\_TKEO\_mean\_dec\_14 & 0.032 \\
        std\_11th\_delta & 0.036 \\
        tqwt\_kurtosisValue\_dec\_20 & 0.004 \\
        app\_det\_TKEO\_mean\_10\_coef & 0.004 \\
        app\_entropy\_shannon\_8\_coef & 0.004 \\
        minIntensity & 0.008 \\
        \hline
    \end{tabular}

\end{document}